\pdfoutput=1

\documentclass[11pt]{article}

\usepackage{emnlp2021}
\usepackage{comment}
\usepackage{times}
\usepackage{latexsym}
\usepackage{graphicx}
\usepackage{color}

\usepackage[T1]{fontenc}

\usepackage[utf8]{inputenc}

\usepackage{microtype}
\usepackage{subfig}
\usepackage{graphicx}
\usepackage{booktabs}
\title{Balancing Methods for Multi-label Text Classification with Long-Tailed Class Distribution}

\usepackage{amsmath}

\author{Yi Huang \\
  Data and Analytics Chapter \\ Roche (China) Holding Ltd., China \\
  \small{\texttt{yi.huang.yh4@roche.com}} \\\And
  Buse Giledereli \\
  Computer Engineering \\ Bogazici University, Turkey \\
  Data and Analytics Chapter \\ Roche Müstahzarları San A.Ş., Turkey \\
  \small{\texttt{busegiledereli@gmail.com}} \\\AND
  Abdullatif Köksal \\
  Computer Engineering \\ Bogazici University, Turkey \\
  \small{\texttt{abdullatif.koksal@boun.edu.tr}} \\\And
  Arzucan Özgür \\
  Computer Engineering \\ Bogazici University, Turkey \\
  \small{\texttt{arzucan.ozgur@boun.edu.tr}} \\\And
  Elif Ozkirimli \\
  Data and Analytics Chapter \\ F. Hoffmann-La Roche AG \\ Switzerland \\
  \small{\texttt{elif.ozkirimli@roche.com}} \\}

\begin{document}
\maketitle
\begin{abstract}

Multi-label text classification is a challenging task because it requires capturing label dependencies. It becomes even more challenging when class distribution is long-tailed. Resampling and re-weighting are common approaches used for addressing the class imbalance problem, however, they are not effective when there is label dependency besides class imbalance because they result in oversampling of common labels. Here, we introduce the application of balancing loss functions 
for multi-label text classification. We perform experiments on a general domain dataset with 90 labels (Reuters-21578) and a domain-specific dataset from PubMed with 18211 labels. 
We find that a distribution-balanced loss function, which inherently addresses both the class imbalance and label linkage problems, outperforms commonly used loss functions. Distribution balancing methods have been successfully used in the image recognition field. Here, we show their effectiveness in natural language processing. Source code is available at \url{https://github.com/Roche/BalancedLossNLP}.

\end{abstract}

\section{Introduction}

Multi-label text classification is one of the core topics in natural language processing (NLP) and is used in many applications such as search  \citep{search_engine} and product categorization \citep{product_cat}. 
It aims to find the related labels from a fixed-set of labels for a given text that may have multiple labels. Figure~\ref{fig:example} demonstrates  examples from the Reuters-21578 multi-label text classification dataset  \citep{Reuters}. Here, for the document with the title \textit{PENN CENTRAL <PC> SELLS U.K. UNIT}, the aim is to find the labels \textit{acq (acquisitions), strategic-metal}, and \textit{nickel} from 90 labels.   
\begin{figure}
    \centering
    \begin{tabular}{ll}
    \toprule
    \textbf{Title 1} &  \begin{tabular}{@{}l@{}l@{}} PENN CENTRAL <PC>  \\ SELLS U.K. UNIT \end{tabular} \\
    \textbf{Labels 1} & \begin{tabular}{@{}l@{}l@{}} acq (1650), strategic-metal (16) \\ \textbf{nickel} (8) \end{tabular} \\
    \midrule
    \textbf{Title 2} & \begin{tabular}{@{}l@{}l@{}} U.S. MINT SEEKING OFFERS \\ ON COPPER, NICKEL \end{tabular} \\ 
    \textbf{Labels 2} & copper (47), \textbf{nickel} (8) \\
    \bottomrule
    \end{tabular}
    \caption{Samples for multi-label text classification task from Reuters-21578 dataset. Only titles are shown for illustration. Numbers after the labels indicate total number of occurrences in the dataset.}
    \label{fig:example}
\end{figure}

Multi-label classification becomes complicated when there is a long-tailed distribution (class imbalance) and linkage (co-occurrence) of labels. Class imbalance occurs when a small subset of the labels (namely head labels) have many instances, while majority of the labels (namely tail labels) have only a few instances. For example, half of the labels in the Reuters dataset, including \textit{copper, strategic-metal,} and \textit{nickel}, occur in less than $5\%$ of the training data. Label co-occurrence or label linkage is a challenge when some head labels co-occur with rare or tail labels, resulting in bias for classification to the head labels. For example, even though the label \textit{nickel} occurs less frequently, the co-occurrence information of \textit{nickel/copper, nickel/strategic-metal} is important for accurate modeling (Figure \ref{fig:example}). Solutions such as resampling of the samples with less-frequent labels in classification \citep{resampling_general_example, resampling_multilabel_example}, using co-occurrence information in the model initialization  \citep{kurata2016improved}, or providing a hybrid solution for head and tail categories with a multi-task architecture \citep{hscnn} have been proposed in NLP, however they are not suitable for imbalanced datasets or they are dependent on the model architecture. 

Multi-label classification has been widely studied in the computer vision (CV) domain, and recently has benefited from cost-sensitive learning through loss functions for tasks such as object recognition \citep{BCE_CV,Dice-Loss-CV}, semantic segmentation \citep{Ge_2018_CVPR}, and medical imaging  \citep{li_multi-label_2020}. Balancing loss functions such as focal loss \citep{lin2017focal}, class-balanced loss \citep{class-balanced-loss} and distribution-balanced loss \citep{DBLoss} provide improvements to resolve the class imbalance and co-occurrence problems in multi-label classification in CV.
Loss function manipulation 
has also been explored   \citep{li-etal-2020-dice, SPECTER} in NLP as it works in a model architecture-agnostic fashion by explicitly embedding the solution into the objective. For example,  \citet{li-etal-2020-dice} has borrowed dice-based loss function from a medical image segmentation task \citep{Dice-Loss-CV} and reported significant improvements over the standard cross-entropy loss function in several NLP tasks. 

In this work, our major contribution is the introduction of the use of 
balancing loss functions to the NLP domain for the multi-label text classification task. We perform experiments on Reuters-21578, a general and small dataset, and PubMed, a biomedical domain-specific and large dataset. For both datasets, the distribution balancing methods not only outperform the other loss functions for the total metrics, but also lead to significant improvement for the tail labels. We suggest that the balancing loss functions provide a robust solution for addressing the challenges in multi-label text classification. 

\section{Loss Functions}
\label{section:loss_functions}

In NLP, Binary Cross Entropy (BCE) loss is commonly used for  multi-label text classification \citep{Bengio-review}. Given a dataset $\{(x^1,y^1),..., (x^N,y^N)\}$ with $N$ training instances, each having a multi-label ground truth of $y^k = [y_{1}^{k},...,y_{C}^{k}] \in \{0,1\}^C$ ($C$ is the number of classes), and a classifier output $z^k = [z_{1}^{k},...,z_{C}^{k}] \in R$, BCE is defined as (the average reduction step is not shown for simplicity):

\small
  \begin{equation}
    L_{BCE}=
    \begin{cases}
      -log(p_{i}^{k}) & \text{if}\ y_{i}^{k}=1 \\
      -log(1-p_{i}^{k}) & \text{otherwise.}
    \end{cases}
  \end{equation}
\normalsize

The $sigmoid$ function is used for computing $p_{i}^{k}$,  $p_{i}^{k} = \sigma(z_{i}^{k})$ . The plain BCE is vulnerable to label imbalance due to the dominance of head classes or negative instances   \citep{BCE_CV}. Below, we describe three alternative approaches that address the class imbalance problem in long-tailed datasets in multi-label text classification. The main idea of these balancing methods is to reweight BCE so that rare instance-label pairs intuitively get reasonable “attention”.

\begin{figure}[t]%
    \centering
    \includegraphics[width=6cm]{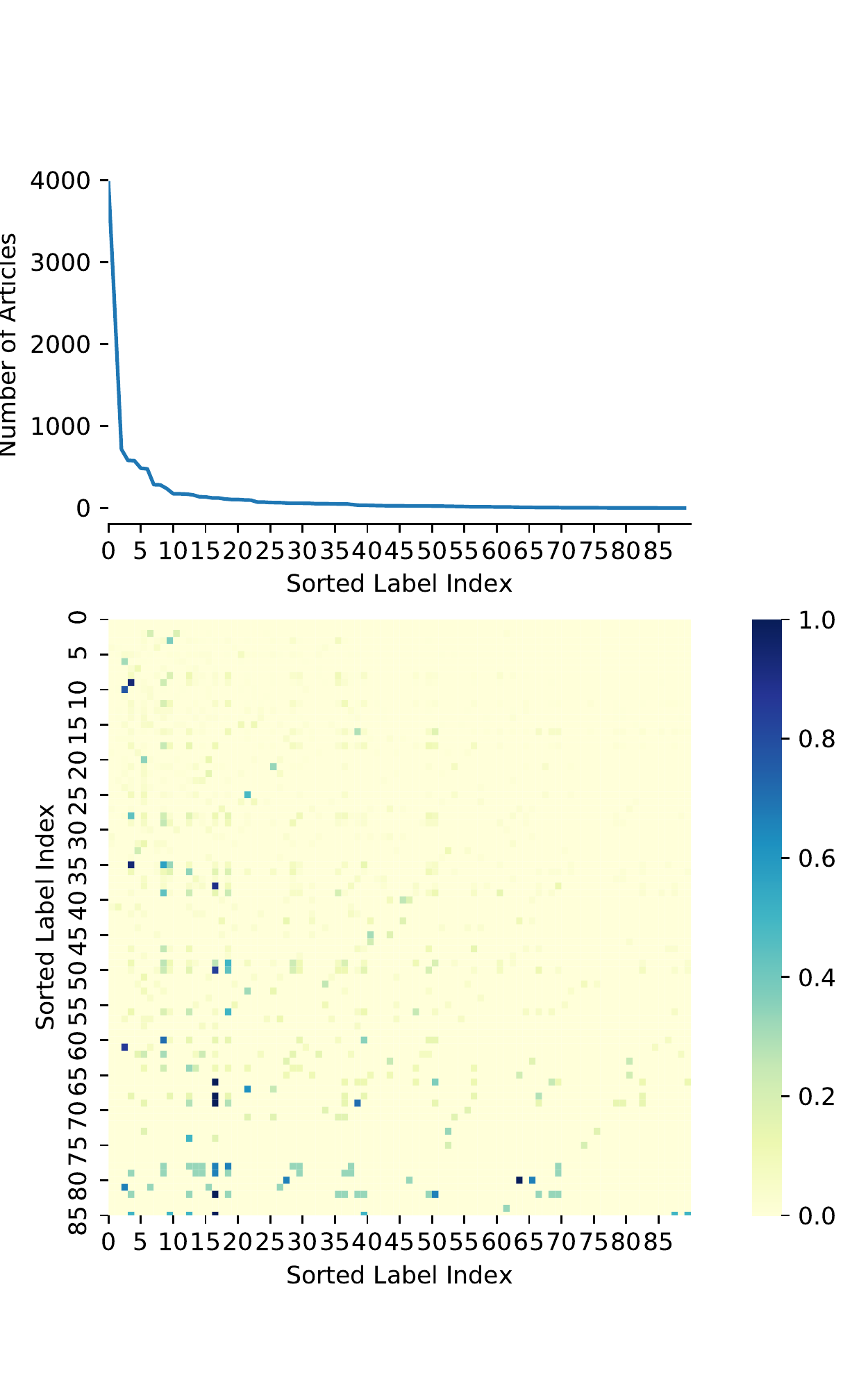}
    \caption{The long-tailed distribution and label co-occurrence for the Reuters-21578 dataset. 
    The co-occurence matrix is color coded based on the conditional probability $p(i|j)$ of class in the $i^{th}$ column on class in the $j^{th}$ row.}%
    \label{fig:longtailed}%
\end{figure}

%

\subsection{Focal loss (FL)}
By multiplying a modulating factor to BCE (with the tunable focusing parameter $\gamma\ge 0$), focal loss places a higher weight of loss on “hard-to-classify” instances predicted with low probability on ground truth \citep{lin2017focal}.
For the multi-label classification task, the focal loss can be defined as:

\small
  \begin{equation}
    L_{FL}=
    \begin{cases}
      -(1-p_{i}^{k})^{\gamma}log(p_{i}^{k}) & \text{if}\ y_{i}^{k}=1 \\
      -{(p_{i}^{k})}^{\gamma}log(1-p_{i}^{k}) & \text{otherwise.}
    \end{cases}
  \end{equation}
\normalsize

\subsection{Class-balanced focal loss (CB)}
By estimating the effective number of samples, class-balanced focal loss \citep{class-balanced-loss} further reweights FL to capture the diminishing marginal benefits of data, and therefore reduces redundant information of head classes.
For multi-label tasks, each label with overall frequency $n_{i}$ has its balancing term

\small
\begin{equation}
r_{CB} = \frac{1-\beta}{1-\beta^{n_i}}
\end{equation}
\normalsize

where 
$\beta \in [0,1)$ controls how fast the effective number grows
and the loss function becomes

\small
  \begin{equation}
    L_{CB}=
    \begin{cases}
      -r_{CB} (1-p_{i}^{k})^{\gamma}log(p_{i}^{k}) & \text{if}\ y_{i}^{k}=1 \\
      -r_{CB} {(p_{i}^{k})}^{\gamma}log(1-p_{i}^{k}) & \text{otherwise.}
    \end{cases}
  \end{equation}
\normalsize

\subsection{Distribution-balanced loss (DB)}
By integrating rebalanced weighting and negative-tolerant regularization (NTR), distribution-balanced loss first reduces redundant information of label co-occurrence, which is critical in the multi-label scenario, and then explicitly assigns lower weight on “easy-to-classify” negative instances \citep{DBLoss}.

First, to rebalance the weights, in the single-label scenario, an instance can be weighted by the resampling probability $P_{i}^{C} = \frac{1}{C} \frac{1}{n_{i}}$; while in the multi-label scenario, if following the same strategy, one instance with multiple labels can be over-sampled with a probability $P^{I} = \frac{1}{C} \sum_{y_i^k=1}\frac{1}{n_{i}}$. Therefore, the rebalanced weight can be normalized with $r_{DB} = P_{i}^{C}/P^{I}$.
With a smoothing function, $\hat{r}_{DB} = \alpha + \sigma(\beta \times (r_{DB} -\mu))$, mapping $r_{DB}$ to $[\alpha, \alpha+1]$, 
the rebalanced-FL (R-FL) loss function is defined as:

\small
  \begin{equation}
    L_{R-FL}=
    \begin{cases}
      -\hat{r}_{DB} (1-p_{i}^{k})^{\gamma}log(p_{i}^{k}) & \text{if}\ y_{i}^{k}=1 \\
      -\hat{r}_{DB} {(p_{i}^{k})}^{\gamma}log(1-p_{i}^{k}) & \text{otherwise.}
    \end{cases}
  \end{equation}
\normalsize

Then, NTR treats the positive and negative instances of the same label differently. A scale factor $\lambda$ and an intrinsic class-specific bias $v_i$ are introduced to lower the threshold for tail classes and to avoid over-suppression. 

\small
  \begin{equation}
    L_{NTR-FL}=
    \begin{cases}
      -(1-q_{i}^{k})^{\gamma}log(q_{i}^{k}) & \text{if}\ y_{i}^{k}=1 \\
      -\frac{1}{\lambda} {(q_{i}^{k})}^{\gamma}log(1-q_{i}^{k}) & \text{otherwise.}
    \end{cases}
  \end{equation}
\normalsize

where $q_{i}^{k}$ = $\sigma(z_{i}^{k}-v_{i})$ for positive instances and $q_{i}^{k}$ = $\sigma(\lambda (z_{i}^{k}-v_{i}))$ for negative ones.
The $v_{i}$ can be estimated by minimizing the loss function at the beginning of training with a scale factor $\kappa$ and class prior $p_{i} = n_{i}/N$, so that 

\small
\begin{equation}
\hat{b}_{i} = -log(\frac{1}{p_i}-1), v_i = -\kappa \times \hat{b}_{i}
\end{equation}
\normalsize

Finally, DB integrates rebalanced weighting and NTR as

\small
  \begin{equation}
    L_{DB}=
    \begin{cases}
      -\hat{r}_{DB}(1-q_{i}^{k})^{\gamma}log(q_{i}^{k}) & \text{if}\ y_{i}^{k}=1 \\
      -\hat{r}_{DB}\frac{1}{\lambda} {(q_{i}^{k})}^{\gamma}log(1-q_{i}^{k}) & \text{otherwise.}
    \end{cases}
  \end{equation}
\normalsize

\section{Experiments}

\subsection{Datasets}
Two multi-label text classification datasets of different size, property and domain are used (Table~\ref{tab:dataset}).

\begin{table}
\caption{Dataset Statistics}
\resizebox{.475\textwidth}{!}{ 
\begin{tabular}{ll}
\toprule
\textbf{Dataset} & \textbf{Statistic} \\
\midrule
\textbf{Reuters-21578} & \\
\ Number of documents & 10788 \\
\ Number of labels & 90  \\
\ Average number of labels per instance                     & 1.24  \\
\ Average number of instances per label                 & 148.11 \\
\midrule 
\textbf{PubMed} & \\
\ Number of documents & 224897 \\
\ Number of labels & 18211  \\
\ Average number of labels per instance                     & 12.30  \\
\ Average number of instances per label                 & 151.88\\
\bottomrule
\end{tabular}
}

\label{tab:dataset}
\end{table}


\textbf{Reuters-21578} dataset (Distribution 1.0) contains documents that appeared on Reuters newswire in 1987 and that were manually annotated with 90 labels  \citep{Reuters}. Here, we follow the train-test split used by \citep{yang-1999} to obtain 7769 training (1000 among which for validation) and 3019 test documents. The labels are equally split into head (30 with $\ge$ 35 instances), medium (31 with between 8-35 instances) and tail (30 with $\le$ 8 instances) subsets.

\begin{table*}
    \small
    \centering
    \caption{Micro and macro F1 scores for multi-label classification of Reuters-21578 (left) and PubMed (right) using the SVM model or different loss functions. The F1 scores are reported for the total set of labels as well as for the head, medium and tail label sets, with the number of instances given in parenthesis. The experiments are performed with the SVM one-vs-rest model (SVM), the binary cross entropy (BCE), focal loss (FL), class balanced focal loss (CB), rebalanced focal loss (R-FL), negative-tolerant regularization FL (NTR-FL), distribution balance with no FL (DB-0FL), class balanced FL with negative regularization (CB-NTR) and distribution balanced loss (DB).
    }
    \setlength\tabcolsep{4pt} 
    \begin{tabular}{lllllllll}
    \toprule
    \begin{tabular}{@{}l@{}}\textbf{Model/} \\ \textbf{Loss} \\ \textbf{Function}  \end{tabular}  & 
    \begin{tabular}{@{}l@{}l@{}}\textbf{Reuters} \\ \textbf{Total} \\ \textbf{miF/maF} \end{tabular}  & 
    \begin{tabular}{@{}l@{}l@{}}\textbf{Reuters} \\ \textbf{Head($\ge$35)}\\ \textbf{miF/maF} \end{tabular}  & 
    \begin{tabular}{@{}l@{}l@{}}\textbf{Reuters} \\ \textbf{Med(8-35)} \\ \textbf{miF/maF} \end{tabular}  & 
    \begin{tabular}{@{}l@{}l@{}}\textbf{Reuters} \\ \textbf{Tail($\le$8)} \\ \textbf{miF/maF} \end{tabular}  & 
    \begin{tabular}{@{}l@{}l@{}}\textbf{PubMed} \\ \textbf{Total} \\ \textbf{miF/maF} \end{tabular}  & 
    \begin{tabular}{@{}l@{}l@{}}\textbf{PubMed} \\ \textbf{Head($\ge$50)} \\ \textbf{miF/maF} \end{tabular}  & 
    \begin{tabular}{@{}l@{}l@{}}\textbf{PubMed} \\ \textbf{Med(15-50)} \\ \textbf{miF/maF} \end{tabular}  & 
    \begin{tabular}{@{}l@{}l@{}}\textbf{PubMed} \\ \textbf{Tail($\le$15)} \\ \textbf{miF/maF} \end{tabular}  \\    
        \midrule
        SVM & 87.60/51.63 & 89.87/78.47 & 66.92/61.00 & 22.54/13.83 & 58.54/13.31 & 60.77/34.33 & 19.78/5.62 & 6.94/0.67 \\
        \midrule
        BCE & 89.14/47.32 & 91.75/82.81 & 66.28/57.26 & 0.00/0.00 & 26.17/0.02 & 27.61/0.06 & 0.00/0.00 & 0.00/0.00 \\
        FL & 89.97/56.83 & 91.83/82.64 & 76.16/70.63 & 27.40/15.37 & 58.30/13.94 & 60.43/33.69 & 26.39/8.15 & 8.58/0.86 \\
        CB & 89.23/52.96 & 91.56/80.44 & 71.64/66.61 & 23.08/9.93 &  58.57/13.67 & 60.75/33.40 & 24.50/7.39 & 9.92/1.01 \\
        R-FL & 89.47/54.35 & 91.59/80.39 & 72.86/66.69 & 25.00/14.22 & 57.90/14.66 & 59.85/34.09 & 30.32/9.70 & 11.45/1.15 \\
        NTR-FL  & 90.70/60.70 & 92.37/82.65 & 79.35/75.34 & 39.51/22.33 &  60.92/16.99 & \textbf{63.15}/38.85 & 33.14/11.39 & 15.86/1.82 \\
        \midrule
        DB-0FL  & 89.45/57.98 & 91.21/82.05 & 77.33/71.11 & 31.17/19.05 &  58.95/15.15 & 60.99/34.92 & 31.06/10.02 & 14.23/1.49\\
        CB-NTR  & \textbf{90.74}/63.31 & \textbf{92.46}/83.28 & 78.42/72.98 & 46.32/\textbf{32.31} &  \textbf{61.07}/18.40 & 63.02/39.95 & 37.18/13.43 & 24.15/2.97 \\
        \midrule
        DB  & 90.62/\textbf{64.47} & 92.14/\textbf{83.48} & \textbf{80.25}/\textbf{77.01} & \textbf{48.89}/31.39 &  60.63/\textbf{19.19} & 62.39/\textbf{40.48} & \textbf{41.14}/\textbf{15.33} & \textbf{24.19}/\textbf{3.08}    \\
    \bottomrule
    \end{tabular}
    \label{tab:results}
\end{table*}

\textbf{PubMed} dataset comes from the BioASQ Challenge  (License Code: 8283NLM123) providing PubMed articles with titles and abstracts, that have been manually labelled for Medical Subject Headings (MeSH)   \citep{BIOASQ, pubmed_nar}. 224,897 articles published during 2020 and 2021 are used, among which 10,000 are used for validation and testing purpose. 
The 18,211 labels are split by 3-quantiles into head (6018 with $\ge$ 50 instances), medium (5581 with between 15-50 instances) and tail (6612 with $\le$ 15 instances) subsets.

\subsection{Experimental Settings}
We compare the use of 
different loss functions, and SVM one-vs-rest model as a classical multi-label classification baseline. 
For each dataset and method, we evaluate its best micro-F1 and macro-F1 scores   \citep{wu-etal-2019-learning-learn, Lipton-F1} for the whole label set (total) as well as different 
subsets of label frequency (head/medium/tail). The loss function parameters, the classification models used, and the implementation details are provided in Appendix \ref{sec:appendix}. 


\section{Results}
A summary of the results of different loss functions are listed in Table \ref{tab:results}.

There are about 10,000 documents and 90 labels in the Reuters dataset, with an average of 150 instances per label (Table \ref{tab:dataset}). Figure \ref{fig:longtailed} shows the long-tailed distribution where only a few labels have a high number of articles and these head labels also have high co-occurrence with other labels.  The impact of the skewed distribution can also be seen from the comparison between the micro-F1 (around 90 for different loss functions) and macro-F1 (around 50-60) scores (Table \ref{tab:results}). Furthermore, 
among loss functions, BCE
has the lowest performance for the Reuters dataset with total macro-F1 score of 47 and tail F1 scores of 0. The PubMed dataset contains around 225,000 documents with 18,000 labels (Table \ref{tab:dataset}) and the imbalance is even more pronounced for this large dataset (Figure in Appendix) and the difference between the total micro-F1 score (60) and the total macro-F1 score (around 15) is very high. 
Overall, SVM underperforms the proposed distribution balanced loss functions in both datasets.

\textbf{Experiments with Reuters-21578 dataset.} 
The loss functions FL, CB, R-FL and NTR-FL perform similar to BCE in head classes, yet outperform BCE in medium and tail classes, indicating the advantage of handling imbalance. DB provides the biggest improvement in tail class assignment; the tail micro-F1 score gains 21.49 from FL and 25.81 from CB.
It outperforms prior works that also used this commonly used dataset, including approaches based on Binary Relevance, EncDec, CNN, CNN-RNN, Optimal Completion Distillation or attention-based GNN, that achieved micro-F1<89.9 \citep{nam_neurips, pal2020magnet, tsai2020order}
 
\textbf{Experiments with PubMed dataset.} 
PubMed is a biomedical domain specific, larger dataset with bigger class imbalance. For this dataset, BCE does not work efficiently, therefore we use FL as a strong baseline. With FL, the medium and tail micro-F1 scores are 26 and 9. All other loss functions outperform FL in medium and tail classes, indicating the advantage of balancing label distribution. DB again has the highest performance for all classes but the most significant improvement is achieved for the medium (micro-F1:41) and tail (micro-F1:24) classes.

\textbf{Ablation Study.}
We further investigate the contribution of the three layers of DB by comparing DB results with 
R-FL, NTR-FL and DB without the focal layer (DB-0FL). As shown in Table \ref{tab:results}, 
for both datasets, removing the NTR layer (R-FL) or the focal layer (DB-0FL) reduces model performance for all subsets. 
Removing the rebalanced weighting layer (NTR-FL) yields similar total micro-F1 (Reuters: 90, PubMed:60) but the macro-F1 as well as medium and tail F1 scores are higher with DB 
, showing the value of adding the rebalancing weighting layer. We also test the contribution of NTR by integrating it with CB, yielding a novel loss function CB-NTR that has not been previously explored. For both datasets, CB-NTR has better performance than CB for all class sets (Table \ref{tab:results}). The only difference between CB-NTR and DB is the use of CB weight ${r}_{CB}$ instead of the rebalancing weight $\hat{r}_{DB}$. DB has very close performance to or outperforms CB-NTR in the medium and tail classes, suggesting that the $\hat{r}_{DB}$ weight, which addresses the co-occurrence challenge, is useful.

\textbf{Error Analysis.}
We perform an error analysis and observe that the most common errors are due to incorrect classification to similar or linked labels for all loss functions. The most common three pairs of classes confused by all loss functions for the Reuters dataset are: \textit{platinum} and \textit{gold}, \textit{yen} and \textit{money-fx}, \textit{platinum} and \textit{copper}. For the PubMed dataset, the most common errors are: \textit{Pandemics} and \textit{Betacoronavirus}, \textit{Pandemics} and \textit{SARS-CoV-2}, \textit{Pneumonia, Viral} and \textit{Betacoronavirus}, and BCE has significantly more errors for these classes compared to the other investigated loss functions.



\section{Conclusion}
We propose and compare the application of a series of balancing loss functions  to address the class imbalance problem in multi-label text classification. We first introduce the loss function DB to NLP and design a novel loss function CB-NTR.
The experiments show that the DB outperforms other approaches by considering long-tailed distribution and label co-occurrence, and its performance is robust to different datasets such as Reuters (90 labels, general domain) and PubMed (18,211 labels, biomedical domain).  This study demonstrates that addressing challenges such as class imbalance and label co-occurrence through loss functions is an effective approach for multi-label text classification.
It does not require additional information 
and can be used with all types of neural network-based models. It
may also be a powerful strategy for other NLP tasks, such as part-of-speech tagging, named entity recognition, machine reading comprehension, paraphrase identification and coreference resolution, all of which usually suffer from long-tailed distribution. 




\section*{Acknowledgements}
We thank Igor Kulev for the helpful discussions, and the anonymous reviewers for their constructive suggestions.
TUBITAK-BIDEB 2211-A Scholarship Program (to A.K.) and TUBA-GEBIP Award of the Turkish Science Academy (to A.O.) are gratefully acknowledged.

\bibliography{anthology,custom}

\begin{thebibliography}{29}
\expandafter\ifx\csname natexlab\endcsname\relax\def\natexlab#1{#1}\fi

\bibitem[{Agrawal et~al.(2013)Agrawal, Gupta, Prabhu, and Varma}]{product_cat}
Rahul Agrawal, Archit Gupta, Yashoteja Prabhu, and Manik Varma. 2013.
\newblock Multi-label learning with millions of labels: Recommending advertiser
  bid phrases for web pages.
\newblock In \emph{Proceedings of the 22nd international conference on World
  Wide Web}, pages 13--24.

\bibitem[{Bengio et~al.(2013)Bengio, Courville, and Vincent}]{Bengio-review}
Yoshua Bengio, Aaron Courville, and Pascal Vincent. 2013.
\newblock \href {https://doi.org/10.1109/TPAMI.2013.50} {Representation
  learning: A review and new perspectives}.
\newblock \emph{IEEE Transactions on Pattern Analysis and Machine
  Intelligence}, 35(8):1798--1828.

\bibitem[{Charte et~al.(2015)Charte, Rivera, del Jesus, and
  Herrera}]{resampling_multilabel_example}
Francisco Charte, Antonio~J Rivera, Mar{\'\i}a~J del Jesus, and Francisco
  Herrera. 2015.
\newblock Addressing imbalance in multilabel classification: Measures and
  random resampling algorithms.
\newblock \emph{Neurocomputing}, 163:3--16.

\bibitem[{Cohan et~al.(2020)Cohan, Feldman, Beltagy, Downey, and
  Weld}]{SPECTER}
Arman Cohan, Sergey Feldman, Iz~Beltagy, Doug Downey, and Daniel Weld. 2020.
\newblock \href {https://doi.org/10.18653/v1/2020.acl-main.207} {{SPECTER}:
  Document-level representation learning using citation-informed transformers}.
\newblock In \emph{Proceedings of the 58th Annual Meeting of the Association
  for Computational Linguistics}, pages 2270--2282, Online. Association for
  Computational Linguistics.

\bibitem[{Coordinators(2017)}]{pubmed_nar}
NCBI~Resource Coordinators. 2017.
\newblock \href {https://doi.org/10.1093/nar/gkx1095} {{Database resources of
  the National Center for Biotechnology Information}}.
\newblock \emph{Nucleic Acids Research}, 46(D1):D8--D13.

\bibitem[{Cui et~al.(2019)Cui, Jia, Lin, Song, and
  Belongie}]{class-balanced-loss}
Yin Cui, Menglin Jia, Tsung-Yi Lin, Yang Song, and Serge Belongie. 2019.
\newblock \href {https://doi.org/10.1109/CVPR.2019.00949} {Class-balanced loss
  based on effective number of samples}.
\newblock In \emph{2019 IEEE/CVF Conference on Computer Vision and Pattern
  Recognition (CVPR)}, pages 9260--9269.

\bibitem[{Devlin et~al.(2018)Devlin, Chang, Lee, and
  Toutanova}]{devlin2018bert}
Jacob Devlin, Ming-Wei Chang, Kenton Lee, and Kristina Toutanova. 2018.
\newblock Bert: Pre-training of deep bidirectional transformers for language
  understanding.
\newblock \emph{arXiv preprint arXiv:1810.04805}.

\bibitem[{Durand et~al.(2019)Durand, Mehrasa, and Mori}]{BCE_CV}
T.~Durand, N.~Mehrasa, and G.~Mori. 2019.
\newblock \href {https://doi.org/10.1109/CVPR.2019.00074} {Learning a deep
  convnet for multi-label classification with partial labels}.
\newblock In \emph{2019 IEEE/CVF Conference on Computer Vision and Pattern
  Recognition (CVPR)}, pages 647--657, Los Alamitos, CA, USA. IEEE Computer
  Society.

\bibitem[{Estabrooks et~al.(2004)Estabrooks, Jo, and
  Japkowicz}]{resampling_general_example}
Andrew Estabrooks, Taeho Jo, and Nathalie Japkowicz. 2004.
\newblock A multiple resampling method for learning from imbalanced data sets.
\newblock \emph{Computational intelligence}, 20(1):18--36.

\bibitem[{Ge et~al.(2018)Ge, Yang, and Yu}]{Ge_2018_CVPR}
Weifeng Ge, Sibei Yang, and Yizhou Yu. 2018.
\newblock Multi-evidence filtering and fusion for multi-label classification,
  object detection and semantic segmentation based on weakly supervised
  learning.
\newblock In \emph{Proceedings of the IEEE Conference on Computer Vision and
  Pattern Recognition (CVPR)}.

\bibitem[{Hayes and Weinstein(1990)}]{Reuters}
Philip~J. Hayes and Steven~P. Weinstein. 1990.
\newblock Construe/tis: A system for content-based indexing of a database of
  news stories.
\newblock In \emph{Proceedings of the The Second Conference on Innovative
  Applications of Artificial Intelligence}, IAAI '90, page 49–64. AAAI Press.

\bibitem[{Kurata et~al.(2016)Kurata, Xiang, and Zhou}]{kurata2016improved}
Gakuto Kurata, Bing Xiang, and Bowen Zhou. 2016.
\newblock Improved neural network-based multi-label classification with better
  initialization leveraging label co-occurrence.
\newblock In \emph{Proceedings of the 2016 Conference of the North American
  Chapter of the Association for Computational Linguistics: Human Language
  Technologies}, pages 521--526.

\bibitem[{Lee et~al.(2019)Lee, Yoon, Kim, Kim, Kim, So, and
  Kang}]{10.1093/bioinformatics/btz682}
Jinhyuk Lee, Wonjin Yoon, Sungdong Kim, Donghyeon Kim, Sunkyu Kim, Chan~Ho So,
  and Jaewoo Kang. 2019.
\newblock \href {https://doi.org/10.1093/bioinformatics/btz682} {{BioBERT: a
  pre-trained biomedical language representation model for biomedical text
  mining}}.
\newblock \emph{Bioinformatics}.

\bibitem[{Li et~al.(2020{\natexlab{a}})Li, Fu, Chen, Li, Liu, Pei, and
  Feng}]{li_multi-label_2020}
Jianqiang Li, Guanghui Fu, Yueda Chen, Pengzhi Li, Bo~Liu, Yan Pei, and Hui
  Feng. 2020{\natexlab{a}}.
\newblock \href {https://doi.org/10.1186/s12859-020-3503-0} {A multi-label
  classification model for full slice brain computerised tomography image}.
\newblock \emph{BMC Bioinformatics}, 21(6):200.

\bibitem[{Li et~al.(2020{\natexlab{b}})Li, Sun, Meng, Liang, Wu, and
  Li}]{li-etal-2020-dice}
Xiaoya Li, Xiaofei Sun, Yuxian Meng, Junjun Liang, Fei Wu, and Jiwei Li.
  2020{\natexlab{b}}.
\newblock \href {https://doi.org/10.18653/v1/2020.acl-main.45} {Dice loss for
  data-imbalanced {NLP} tasks}.
\newblock In \emph{Proceedings of the 58th Annual Meeting of the Association
  for Computational Linguistics}, pages 465--476, Online. Association for
  Computational Linguistics.

\bibitem[{Lin et~al.(2017)Lin, Goyal, Girshick, He, and Dollár}]{lin2017focal}
Tsung-Yi Lin, Priya Goyal, Ross Girshick, Kaiming He, and Piotr Dollár. 2017.
\newblock \href {https://doi.org/10.1109/ICCV.2017.324} {Focal loss for dense
  object detection}.
\newblock In \emph{2017 IEEE International Conference on Computer Vision
  (ICCV)}, pages 2999--3007, Los Alamitos, CA, USA. IEEE Computer Society.

\bibitem[{Lipton et~al.(2014)Lipton, Elkan, and Naryanaswamy}]{Lipton-F1}
Zachary~C. Lipton, Charles Elkan, and Balakrishnan Naryanaswamy. 2014.
\newblock Optimal thresholding of classifiers to maximize f1 measure.
\newblock In \emph{Machine Learning and Knowledge Discovery in Databases},
  pages 225--239, Berlin, Heidelberg. Springer Berlin Heidelberg.

\bibitem[{Milletari et~al.(2016)Milletari, Navab, and Ahmadi}]{Dice-Loss-CV}
Fausto Milletari, Nassir Navab, and Seyed-Ahmad Ahmadi. 2016.
\newblock \href {https://doi.org/10.1109/3DV.2016.79} {V-net: Fully
  convolutional neural networks for volumetric medical image segmentation}.
\newblock In \emph{2016 Fourth International Conference on 3D Vision (3DV)},
  pages 565--571.

\bibitem[{Nam et~al.(2017)Nam, Loza~Menc\'{\i}a, Kim, and
  F\"{u}rnkranz}]{nam_neurips}
Jinseok Nam, Eneldo Loza~Menc\'{\i}a, Hyunwoo~J Kim, and Johannes
  F\"{u}rnkranz. 2017.
\newblock \href
  {https://proceedings.neurips.cc/paper/2017/file/2eb5657d37f474e4c4cf01e4882b8962-Paper.pdf}
  {Maximizing subset accuracy with recurrent neural networks in multi-label
  classification}.
\newblock In \emph{Advances in Neural Information Processing Systems},
  volume~30. Curran Associates, Inc.

\bibitem[{Pal et~al.(2020)Pal, Selvakumar, and Sankarasubbu}]{pal2020magnet}
Ankit Pal, Muru Selvakumar, and Malaikannan Sankarasubbu. 2020.
\newblock Magnet: Multi-label text classification using attention-based graph
  neural network.
\newblock In \emph{ICAART (2)}, pages 494--505.

\bibitem[{Pedregosa et~al.(2011)Pedregosa, Varoquaux, Gramfort, Michel,
  Thirion, Grisel, Blondel, Prettenhofer, Weiss, Dubourg, Vanderplas, Passos,
  Cournapeau, Brucher, Perrot, and Duchesnay}]{scikit-learn}
F.~Pedregosa, G.~Varoquaux, A.~Gramfort, V.~Michel, B.~Thirion, O.~Grisel,
  M.~Blondel, P.~Prettenhofer, R.~Weiss, V.~Dubourg, J.~Vanderplas, A.~Passos,
  D.~Cournapeau, M.~Brucher, M.~Perrot, and E.~Duchesnay. 2011.
\newblock Scikit-learn: Machine learning in {P}ython.
\newblock \emph{Journal of Machine Learning Research}, 12:2825--2830.

\bibitem[{Prabhu et~al.(2018)Prabhu, Kag, Harsola, Agrawal, and
  Varma}]{search_engine}
Yashoteja Prabhu, Anil Kag, Shrutendra Harsola, Rahul Agrawal, and Manik Varma.
  2018.
\newblock Parabel: Partitioned label trees for extreme classification with
  application to dynamic search advertising.
\newblock In \emph{Proceedings of the 2018 World Wide Web Conference}, pages
  993--1002.

\bibitem[{Tsai and Lee(2020)}]{tsai2020order}
Che{-}Ping Tsai and Hung{-}yi Lee. 2020.
\newblock Order-free learning alleviating exposure bias in multi-label
  classification.
\newblock In \emph{The Thirty-Fourth {AAAI} Conference on Artificial
  Intelligence, {AAAI} 2020, The Thirty-Second Innovative Applications of
  Artificial Intelligence Conference, {IAAI} 2020, The Tenth {AAAI} Symposium
  on Educational Advances in Artificial Intelligence, {EAAI} 2020, New York,
  NY, USA, February 7-12, 2020}, pages 6038--6045. {AAAI} Press.

\bibitem[{Tsatsaronis et~al.(2015)Tsatsaronis, Balikas, Malakasiotis, Partalas,
  Zschunke, Alvers, Weissenborn, Krithara, Petridis, Polychronopoulos,
  Almirantis, Pavlopoulos, Baskiotis, Gallinari, Artieres, Ngonga, Heino,
  Gaussier, Barrio-Alvers, Schroeder, Androutsopoulos, and Paliouras}]{BIOASQ}
George Tsatsaronis, Georgios Balikas, Prodromos Malakasiotis, Ioannis Partalas,
  Matthias Zschunke, Michael~R Alvers, Dirk Weissenborn, Anastasia Krithara,
  Sergios Petridis, Dimitris Polychronopoulos, Yannis Almirantis, John
  Pavlopoulos, Nicolas Baskiotis, Patrick Gallinari, Thierry Artieres, Axel
  Ngonga, Norman Heino, Eric Gaussier, Liliana Barrio-Alvers, Michael
  Schroeder, Ion Androutsopoulos, and Georgios Paliouras. 2015.
\newblock \href {https://doi.org/10.1186/s12859-015-0564-6} {An overview of the
  bioasq large-scale biomedical semantic indexing and question answering
  competition}.
\newblock \emph{BMC Bioinformatics}, 16:138.

\bibitem[{Wolf et~al.(2020)Wolf, Debut, Sanh, Chaumond, Delangue, Moi, Cistac,
  Rault, Louf, Funtowicz, Davison, Shleifer, von Platen, Ma, Jernite, Plu, Xu,
  Scao, Gugger, Drame, Lhoest, and Rush}]{wolf-etal-2020-transformers}
Thomas Wolf, Lysandre Debut, Victor Sanh, Julien Chaumond, Clement Delangue,
  Anthony Moi, Pierric Cistac, Tim Rault, Rémi Louf, Morgan Funtowicz, Joe
  Davison, Sam Shleifer, Patrick von Platen, Clara Ma, Yacine Jernite, Julien
  Plu, Canwen Xu, Teven~Le Scao, Sylvain Gugger, Mariama Drame, Quentin Lhoest,
  and Alexander~M. Rush. 2020.
\newblock \href {https://www.aclweb.org/anthology/2020.emnlp-demos.6}
  {Transformers: State-of-the-art natural language processing}.
\newblock In \emph{Proceedings of the 2020 Conference on Empirical Methods in
  Natural Language Processing: System Demonstrations}, pages 38--45, Online.
  Association for Computational Linguistics.

\bibitem[{Wu et~al.(2019)Wu, Xiong, and Wang}]{wu-etal-2019-learning-learn}
Jiawei Wu, Wenhan Xiong, and William~Yang Wang. 2019.
\newblock \href {https://doi.org/10.18653/v1/D19-1444} {Learning to learn and
  predict: A meta-learning approach for multi-label classification}.
\newblock In \emph{Proceedings of the 2019 Conference on Empirical Methods in
  Natural Language Processing and the 9th International Joint Conference on
  Natural Language Processing (EMNLP-IJCNLP)}, pages 4354--4364, Hong Kong,
  China. Association for Computational Linguistics.

\bibitem[{Wu et~al.(2020)Wu, Huang, Liu, Wang, and Lin}]{DBLoss}
Tong Wu, Qingqiu Huang, Ziwei Liu, Yu~Wang, and Dahua Lin. 2020.
\newblock \href {https://doi.org/10.1007/978-3-030-58548-8_10}
  {Distribution-balanced loss for multi-label classification in long-tailed
  datasets}.
\newblock In \emph{Computer Vision -- ECCV 2020}, pages 162--178, Cham.
  Springer International Publishing.

\bibitem[{Yang et~al.(2020)Yang, Li, Fukumoto, and Ye}]{hscnn}
Wenshuo Yang, Jiyi Li, Fumiyo Fukumoto, and Yanming Ye. 2020.
\newblock \href {https://doi.org/10.18653/v1/2020.emnlp-main.545} {{HSCNN}: A
  hybrid-{S}iamese convolutional neural network for extremely imbalanced
  multi-label text classification}.
\newblock In \emph{Proceedings of the 2020 Conference on Empirical Methods in
  Natural Language Processing (EMNLP)}, pages 6716--6722, Online. Association
  for Computational Linguistics.

\bibitem[{Yang and Liu(1999)}]{yang-1999}
Yiming Yang and Xin Liu. 1999.
\newblock \href {https://doi.org/10.1145/312624.312647} {A re-examination of
  text categorization methods}.
\newblock In \emph{Proceedings of the 22nd Annual International ACM SIGIR
  Conference on Research and Development in Information Retrieval}, SIGIR '99,
  page 42–49, New York, NY, USA. Association for Computing Machinery.

\end{thebibliography}
\bibliographystyle{acl_natbib}

\clearpage
\appendix
\section{Appendix}
\label{sec:appendix}

\subsection{Experimental Settings}
\textbf{Evaluation metrics.} For each dataset and method, we select the threshold with the best micro-F1 score on the validation set as our final model and evaluate its performance on the test set with micro-F1 and macro-F1 scores.

\textbf{Loss function parameters.} 
We compare the performance of DB with different loss functions, 
where BCE or its modifications are used.
The methods include: (1) BCE with all instances and labels of the same weight. (2) FL \citep{lin2017focal}: we use $\gamma$=2. (3) CB \citep{class-balanced-loss}: we use $\beta$ =0.9. (4) R-FL \citep{DBLoss}: we use $\alpha$=0.1 and $\beta$=10, $\mu$=0.9 (Reuters-21578) or 0.05 (PubMed). (5)NTR-FL \citep{DBLoss}: we use $\kappa$=0.05 and $\lambda$=2. (6) DB \citep{DBLoss}: we use same parameters with R-FL and NTR-FL when applicable.

\textbf{Implementation Details.} We use the \textit{BertForSequenceClassification} backbone in \textit{transformers} library \citep{wolf-etal-2020-transformers} with the bert-base-cased pretrained model  \citep{devlin2018bert} for Reuters-21578 dataset and the biobert-base-cased-v1.1 pretrained model  \citep{10.1093/bioinformatics/btz682} for PubMed dataset. bert-base-cased and biobert-base-cased-v1.1 are base BERT models with 110 million parameters. The training data are truncated with a maximal length of 512 and grouped with a batch size of 32. We use AdamW with a weight decay of 0.01 as the optimizer, and determine the learning rate by hyperparameter search. 
The experiments are implemented in PyTorch. For Reuters-21578 dataset we use one-GPU (V100) experiments which takes 5 minutes for one epoch. For PubMed dataset, we use one-GPU (A100) experiments which takes 1 hour for one epoch. For the SVM one-vs-rest model, we use \textit{scikit-learn} library \citep{scikit-learn} with TF-IDF features. With hyperparameter search, we apply the linear kernel and hyper-plane shifting optimized on each validation set.

\subsection{Additional Effectiveness Check} 
We further investigate the effectiveness of loss functions against the number of labels per instance (Table \ref{tab:supple-results} in Appendix). For the Reuters dataset, we split the test instances into two groups, 2583 instances with only one label and 436 instances with multiple labels. On single-label instances, all functions from BCE to DB, have similar performance; while on multi-label instances, the performance of BCE drops more than DB. DB outperforms other functions in micro-F1 of the multi-label instance group and macro-F1 of both groups. There are < 0.1\% instances of PubMed dataset with a single label, so we divide instances into 3-quantiles by their number of labels. In each quantile, the novel NTR-FL, CB-NTR and DB outperform the rest of the models in all metrics.


\begin{figure}[t]%
    \centering
    \includegraphics[width=7cm]{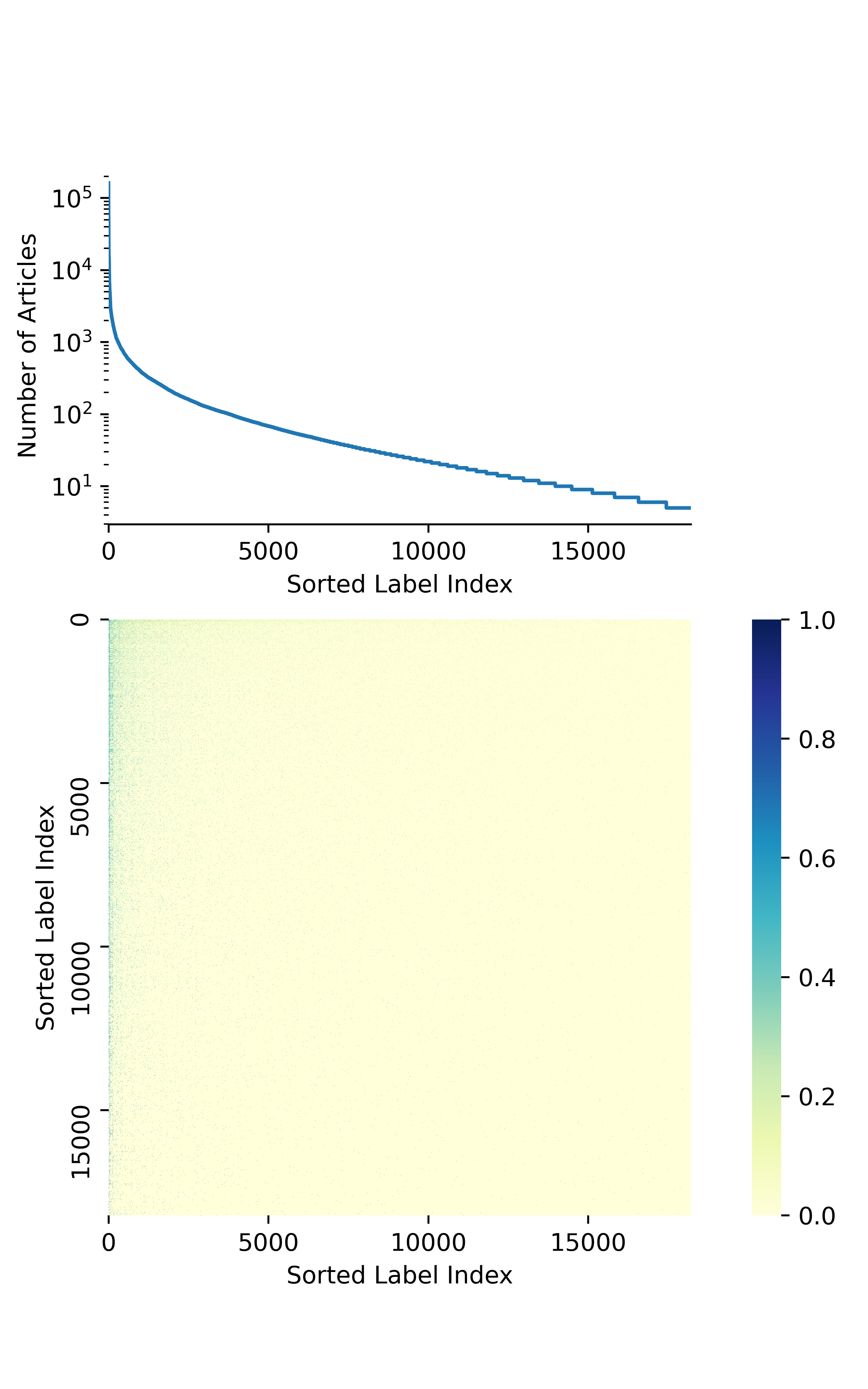}
    \caption{The long-tailed distribution and label co-occurrence for the PubMed dataset. 
    The y-axis of distribution curve is log-scale, and the co-occurence matrix is color coded based on the quad root (for better visualization) of conditional probability $p(i|j)$ of class in the $i^{th}$ column on class in the $j^{th}$ row.}%
    \label{fig:longtailed_pubmed}%
\end{figure}

\begin{table*}
    \small
    \centering
    \caption{Micro and macro F1 scores for multi-label classification of Reuters-21578 (left) and PubMed (right) using different loss functions. The F1 scores are reported for the total set of labels as well as for groups split by the number of labels per instance. The experiments are performed with the binary cross entropy (BCE), focal loss (FL), class balanced focal loss (CB), rebalanced focal loss (R-FL), negative-tolerant regularization FL (NTR-FL), distribution balance with no FL (DB-0FL), class balanced FL with negative regularization (CB-NTR) and distribution balanced loss (DB).
    }
    \setlength\tabcolsep{4pt} 
    \begin{tabular}{llllllll}
    \toprule
    \begin{tabular}{@{}l@{}} \textbf{Loss} \\ \textbf{Function}  \end{tabular}  & 
    \begin{tabular}{@{}l@{}l@{}}\textbf{Reuters} \\ \textbf{Total} \\ \textbf{miF/maF} \end{tabular}  & 
    \begin{tabular}{@{}l@{}l@{}}\textbf{Reuters} \\ \textbf{Single-label}\\ \textbf{miF/maF} \end{tabular}  & 
    \begin{tabular}{@{}l@{}l@{}}\textbf{Reuters} \\ \textbf{Multi-label} \\ \textbf{miF/maF} \end{tabular}  & 
    \begin{tabular}{@{}l@{}l@{}}\textbf{PubMed} \\ \textbf{Total} \\ \textbf{miF/maF} \end{tabular}  & 
    \begin{tabular}{@{}l@{}l@{}}\textbf{PubMed} \\ \textbf{$\le$ 9 labels} \\ \textbf{miF/maF} \end{tabular}  & 
    \begin{tabular}{@{}l@{}l@{}}\textbf{PubMed} \\ \textbf{10-14 labels} \\ \textbf{miF/maF} \end{tabular}  & 
    \begin{tabular}{@{}l@{}l@{}}\textbf{PubMed} \\ \textbf{$\ge$ 15 labels} \\ \textbf{miF/maF} \end{tabular}  \\    
        \midrule
        BCE & 89.14/47.32 & 94.11/41.44 & 76.26/33.11 &  26.17/0.02 & 16.48/0.01 & 27.36/0.02 & 30.36/0.03 \\
        FL & 89.97/56.83 & 94.81/50.33 & 77.54/40.07 &  58.30/13.94 & 53.72/7.44 & 59.02/10.27 & 59.72/8.63 \\
        CB & 89.23/52.96 & 94.10/44.72 & 77.27/38.80 &  58.57/13.67 & 54.41/7.40 & 59.21/10.11 & 59.82/8.51 \\
        R-FL & 89.47/54.35 & 95.21/47.45 & 74.29/38.79 & 57.90/14.66 & 53.08/7.67 & 58.60/10.50 & 59.45/8.81 \\
        NTR-FL  & 90.70/60.70 & \textbf{95.42}/51.33 & 78.85/44.37 &  60.92/16.99 & \textbf{58.51}/9.07 & \textbf{61.86}/12.31 & 61.12/10.20 \\
        \midrule
        DB-0FL  & 89.45/57.98 & 94.48/51.80 & 76.63/42.26 &  58.95/15.15 & 55.14/8.11 & 59.84/10.90 & 59.85/8.94\\
        CB-NTR  & \textbf{90.74}/63.31 & 95.17/51.08 & 79.56/49.94 &  \textbf{61.07}/18.40 & 58.29/9.67 & 61.72/12.97 & \textbf{61.72}/10.77 \\
        \midrule
        DB  & 90.62/\textbf{64.47} & 94.49/\textbf{54.31} & \textbf{81.17}/\textbf{50.12} & 60.63/\textbf{19.19}  & 57.81/\textbf{9.76} & 61.53/\textbf{13.49} & 61.08/\textbf{11.23}    \\
    \bottomrule
    \end{tabular}
    \label{tab:supple-results}
\end{table*}

\end{document}